\documentclass[runningheads]{llncs}

\usepackage{eccv}

\usepackage{eccvabbrv}

\usepackage{graphicx}
\usepackage{booktabs}
\usepackage{svg}
\usepackage{wrapfig}
\usepackage{algorithm}
\usepackage{algorithmic}

\usepackage{listings}
\usepackage{multicol}
\usepackage{sourcecodepro}
\usepackage{mdframed}
\usepackage{longtable}

\definecolor{codegreen}{rgb}{0,0.6,0}
\definecolor{codegray}{rgb}{0.5,0.5,0.5}
\definecolor{codepurple}{rgb}{0.58,0,0.82}
\definecolor{backcolour}{rgb}{1,1,1}
\definecolor{lightgray}{gray}{0.9}
\newcommand{\customsize}{\fontsize{6}{6}\selectfont}

\lstdefinestyle{mystyle}{
    backgroundcolor=\color{backcolour},   
    commentstyle=\color{codegreen},
    keywordstyle=\color{magenta},
    numberstyle=\tiny\color{codegray},
    stringstyle=\color{codepurple},
    basicstyle=\ttfamily\fontseries{l}\customsize,
    breakatwhitespace=false,         
    breaklines=true,                 
    captionpos=b,                    
    keepspaces=true,                 
    numbers=left,                    
    numbersep=5pt,                  
    showspaces=false,                
    showstringspaces=false,
    showtabs=false,                  
    tabsize=2
}
\definecolor{pastelGreen}{RGB}{204, 255, 204}
\definecolor{pastelViolet}{RGB}{216, 191, 216}
\definecolor{pastelRed}{RGB}{255, 204, 204}
\definecolor{highlightGreen}{RGB}{64, 135, 48}
\definecolor{highlightViolet}{RGB}{216, 191, 216}
\definecolor{highlightRed}{RGB}{196, 53, 53}

\usepackage[accsupp]{axessibility}  % Improves PDF readability for those with disabilities.

\usepackage[pagebackref,breaklinks,colorlinks,citecolor=eccvblue]{hyperref}
\usepackage{hyperref}

\usepackage{orcidlink}

\begin{document}

\title{Improving Medical Multi-modal Contrastive Learning with Expert Annotations} 

\titlerunning{eCLIP}

\author{Yogesh Kumar\orcidlink{0000-0002-7961-8596} \and
Pekka Marttinen\orcidlink{0000-0001-7078-7927}}

\authorrunning{Y.~Kumar and P.~Marttinen}

\institute{Department of Computer Science, Aalto University, Finland \\
\email{\{firstname.lastname\}@aalto.fi}}

\maketitle

\begin{abstract}
We introduce eCLIP, an enhanced version of the CLIP model that integrates expert annotations in the form of radiologist eye-gaze heatmaps. It tackles key challenges in contrastive multi-modal medical imaging analysis, notably data scarcity and the ``modality gap'' -- a significant disparity between image and text embeddings that diminishes the quality of representations and hampers cross-modal interoperability. eCLIP integrates a heatmap processor and leverages mixup augmentation to efficiently utilize the scarce expert annotations, thus boosting the model's learning effectiveness. eCLIP is designed to be generally applicable to any variant of CLIP without requiring any modifications of the core architecture. Through detailed evaluations across several tasks, including zero-shot inference, linear probing, cross-modal retrieval, and Retrieval Augmented Generation (RAG) of radiology reports using a frozen Large Language Model, eCLIP showcases consistent improvements in embedding quality. The outcomes reveal enhanced alignment and uniformity, affirming eCLIP's capability to harness high-quality annotations for enriched multi-modal analysis in the medical imaging domain.

  \keywords{Contrastive Learning \and Medical Imaging \and Zero-shot Inference}
\end{abstract}

\section{Introduction}
\label{sec:intro}

Pretraining foundation models on multi-modal data -- particularly leveraging the relationships between text and images -- has proven to be a robust strategy for generating versatile embeddings \cite{radford2021learning,jia2021scaling}. These embeddings enhance the efficacy in several downstream tasks, from image generation to advanced vision-language integration \cite{liu2023visual, ramesh2021zero, singh2022flava}. Central to this approach is the employment of a contrastive learning (CL) loss objective \cite{oord2018representation, chen2020simple, zbontar2021barlow}, where models are trained to align positive pairs (e.g., an image and its corresponding caption) while diversifying negative ones. A significant hurdle in this approach is the necessity of vast datasets, often comprising several millions of data points, for competitive results. Models such as CLIP \cite{radford2021learning} have been trained on internet-scale datasets, estimated to encompass hundreds of millions of image-text pairs \cite{cherti2023reproducible, xu2023demystifying}. Acquiring datasets of this magnitude poses substantial challenges in specialized fields that require expert knowledge for data collection, processing and annotation. The medical imaging domain exemplifies these difficulties, where acquiring even a single data point, such as a chest X-ray, involves complex processes requiring expertise and significant resources. Moreover, the procurement of such data for machine learning research is further complicated by ethical considerations, patient privacy concerns and the need for extensive de-identification procedures. 

\begin{figure}[tb]
  \centering
  \begin{subfigure}{0.6\textwidth}
    \centering
    \includegraphics[height=4cm]{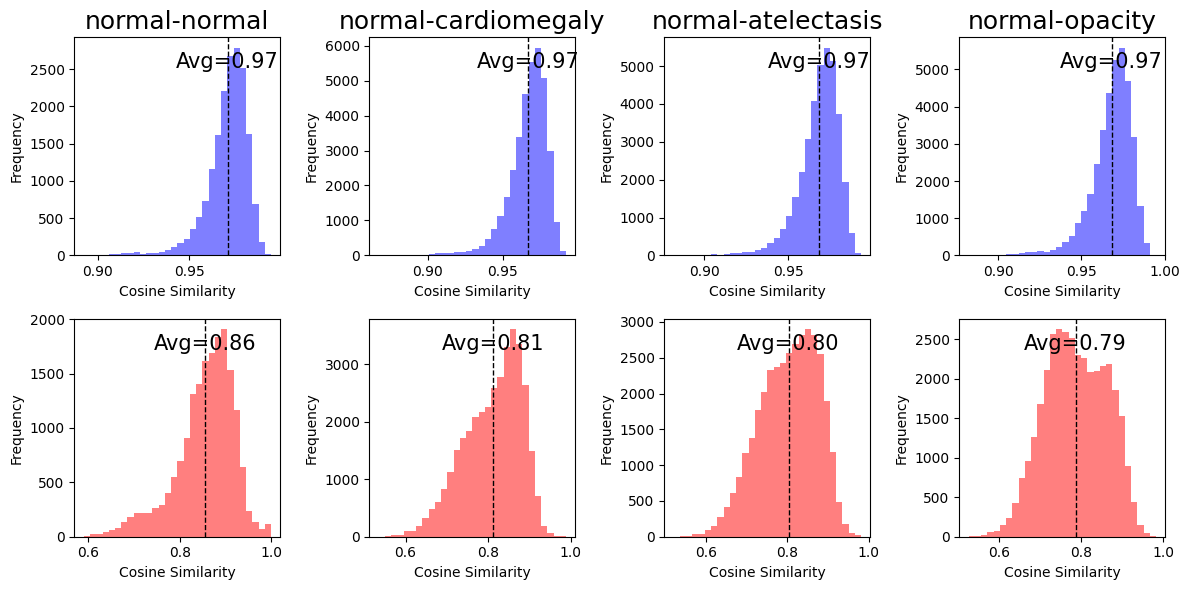}
    \caption{\textbf{Histogram of Cosine Similarities Among Subgroups.} The model exhibits high cosine similarity among embeddings within the text (\emph{red}) and image (\emph{blue}) modalities, regardless of the differences among subgroups. This underscores the model's challenge in capturing the subtleties inherent in medical data.}
    \label{subfig:clip_cosine_hist}
  \end{subfigure}
  \hfill 
  \begin{subfigure}{0.3\textwidth}
    \centering
    \includegraphics[height=3.5cm,width=3.5cm]{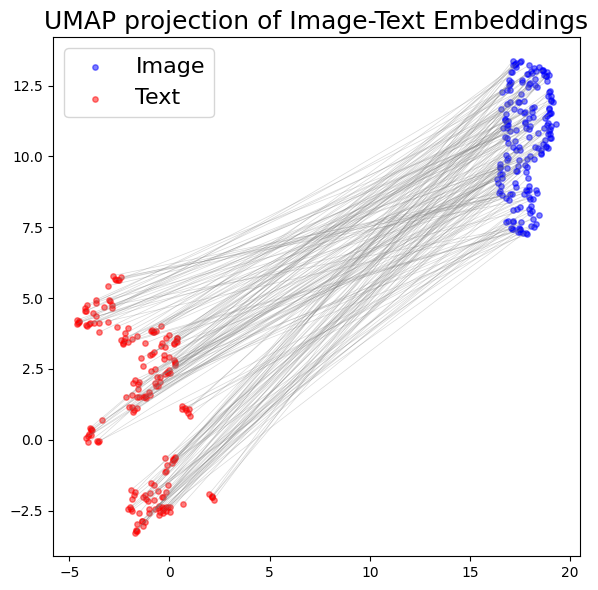} 
    \caption{\textbf{Modality Gap.} Despite the CLIP contrastive loss aiming at closely aligning image and text embeddings within a shared space, the modalities remain segregated into distinct regions.}
    \label{subfig:clip_modality_gap}
  \end{subfigure}
  \caption{\textbf{Analysis of CLIP Embeddings in Medical Imaging} The figure presents embeddings generated by a CLIP model, pretrained on an internet-scale dataset, applied to the Open-I dataset pairing X-rays with corresponding radiology reports.}
  \label{fig:clip-cosine-similarities}
\end{figure}

This has led to the prevalent use of foundation models, initialized with weights from models trained on extensive internet-scale datasets, for tasks in the medical domain \cite{zhang2022contrastive, wang2022medclip, huang2021gloria, krishnan2022self}. However, the areas of interest within medical images are often nuanced and require expert knowledge to interpret, rendering them indistinguishable to a general-purpose model. In \cref{subfig:clip_cosine_hist}, we investigate the embeddings generated by a CLIP model -- initially pretrained on internet data -- using samples from the Open-I dataset\cite{demner2016preparing}, which includes X-rays and corresponding radiology reports. We categorize the samples into subgroups based on the primary abnormality identified in each report, such as `normal', cardiomegaly, atelectasis and opacity. A histogram of the cosine similarities between embeddings from different groups indicates a high degree of similarity, with values approaching 1. This could lead to potential challenges in downstream zero-shot inference tasks, which rely on the spatial segregation of embeddings from different groups \cite{radford2021learning}. Typically, continual pretraining on medically relevant data is employed to enhance the model's ability to differentiate between various abnormalities.

Recent studies \cite{liang2022mind, goel2022cyclip, oh2024geodesic, zhang2023connect, tschannen2023clippo} have identified a ``modality gap'' in multi-modal contrastive representation learning, where the embeddings from different modalities (e.g., images and text) fall in distinct regions in the shared embedding space. This separation, which arises from factors such as initial model weights and the objectives of contrastive learning \cite{liang2022mind, zhang2023connect}, leads to the ``cone effect'' where embeddings of each modality are restricted to a narrow region of the embedding hypersphere. In \cref{subfig:clip_modality_gap}, we illustrate this within the medical domain with a 2D UMAP\cite{mcinnes2018umap} projection of image and text embeddings. This example highlights how embeddings from the same modality but different semantic groups, such as X-ray images of varying abnormalities, cluster closely together. This makes it difficult for a model to distinguish between semantically different images, undermining its performance in medical image analysis.

\begin{figure}[tb]
  \centering
  \begin{minipage}{0.65\textwidth}
    \centering
    \includegraphics[width=\linewidth]{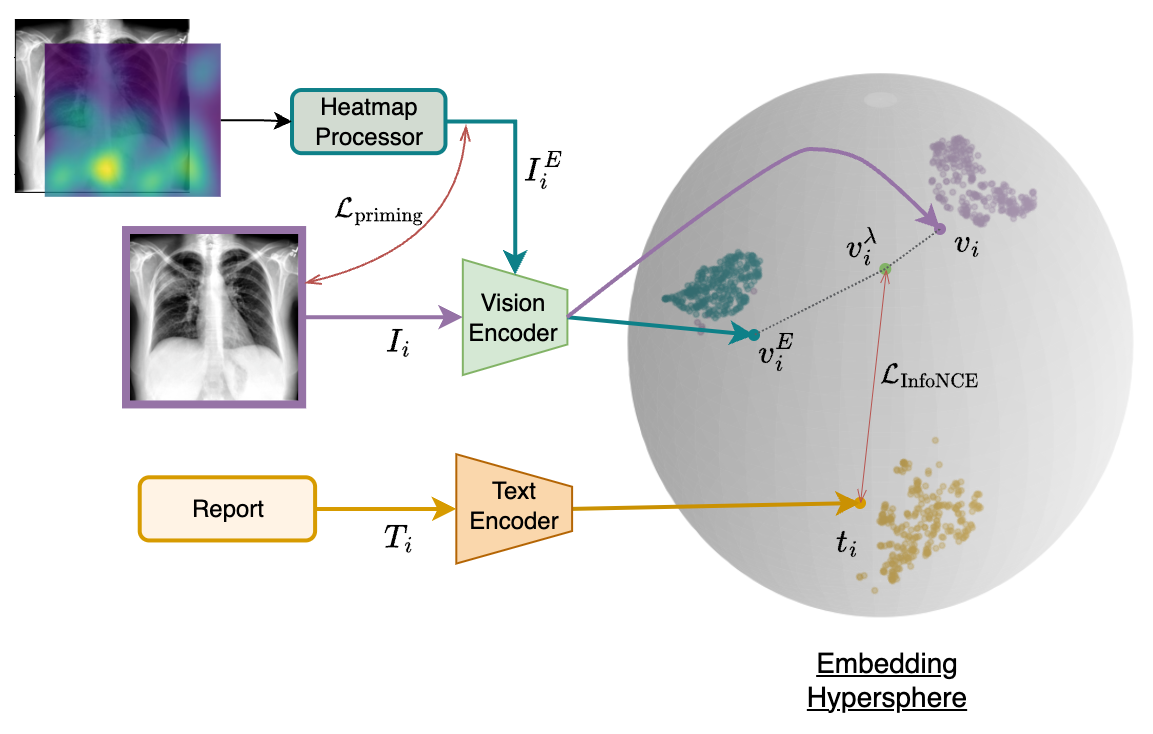}
  \end{minipage}
  \begin{minipage}{0.3\textwidth}
    \centering
    \includegraphics[width=\linewidth]{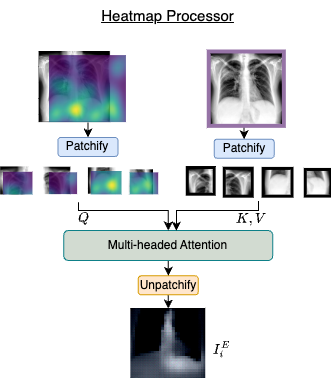}
  \end{minipage}
  \caption{\textbf{eCLIP Pretraining with Expert Annotations.} eCLIP adds a Heatmap Processor \textit{(right)}, featuring a multi-headed attention layer, to the standard Image and Text encoders in CLIP. This processor, along with vision and text encoders, maps inputs into a shared hypersphere. Here, the original image ($I_i$), its text ($T_i$) and the heatmap-processed image ($I^E_i$) are positioned within a tripartite area (shown here after 2D UMAP projection, please refer to the Supplement for a scaled version). We employ mixup between $I_i$ and $I^E_i$ to generate the embedding $v^{\lambda}_i$, which gives us additional positive pairs to enhance the CLIP InfoNCE loss optimization. An auxillary loss, $\mathcal{L}_{\text{priming}}$, is used during the initial training steps to ``prime'' the heatmap processor to imitate an identity function when the heatmap is composed of all ones.}
  \label{fig:main-figure}
\end{figure}

We investigate the potential of integrating expert annotations, specifically radiologist eye-gaze heatmaps, to alleviate these issues. Processing the eye-gaze data from radiologists\cite{karargyris2021creation} provides us with heatmaps indicative of the radiologist's attention across different regions of the X-ray images. This heatmap reflects areas of clinical interest aligned with details present in radiology reports. We posit that this could help capture nuanced visual cues in the X-rays and therefore pairing it with reports can enrich the CLIP training data with high-quality positive pairs. Due to the scarcity of such expert annotated data, we employ the mixup strategy, a data augmentation technique which has been effective in both supervised \cite{zhang2017mixup,verma2019manifold,han2022umix} and contrastive learning\cite{verma2021towards,oh2024geodesic}, to create additional synthetic samples.

We present eCLIP (expert-annotated CLIP), an adaptation of the CLIP model that incorporates expert eye-gaze heatmaps, without modifying the CLIP model's core architecture. The operational workflow of eCLIP is depicted in \cref{fig:main-figure}. Our \textbf{contributions} are as follows:
\begin{itemize}
    \item \textbf{Utilization of Expert Annotations.} We harness radiologist eye-gaze heatmaps to create additional embeddings, effectively introducing valuable positive and negative pairs for enhancing the contrastive learning process. 
    \item \textbf{eCLIP Architecture.} Our implementation features a heatmap processing mechanism utilizing multi-headed attention (MHA), optimized for handling both heatmaps and original images. This is complemented by a mixup strategy to addresses the challenge of data scarcity, and curriculum learning to ensure a gradual introduction of expert annotations.
    \item \textbf{Comprehensive Evaluation.} We assess eCLIP's zero-shot classification accuracy, sample efficiency and cross-modal retrieval performance and embedding quality across multiple chest X-ray datasets. We also evaluate the cross-modal embeddings to generate radiology reports using a frozen Large Language Model (LLM) without explicitly fine-tuning on medical data.
\end{itemize}

\section{Related Work}

\noindent \textbf{Modality Gap: } Liang \etal \cite{liang2022mind} pinpoint the origins of the modality gap to the nuances of model initialization and the objectives of contrastive learning, underscoring its impact on downstream tasks and fairness. Oh \etal \cite{oh2024geodesic} highlight poor uniformity and alignment in CLIP's embeddings and propose a finetuning method for robust representations. Zhang \etal \cite{zhang2023connect} explore the geometry of this embedding space, and provide both theoretical and empirical insights on the nature of this geometry. Subsequent research has produced methods to mitigate the modality gap through diverse and creative approaches \cite{goel2022cyclip, tschannen2023clippo, zhang2023diagnosing, gu2023can, nukrai2022text}.

\noindent \textbf{Improving Contrastive Learning: } 
Several methods have improved upon the CLIP objective by introducing auxilliary losses, e.g., SLIP \cite{mu2022slip} uses SimCLR \cite{chen2020simple} loss, M3MAE \cite{geng2205m3ae, weers2023masked} augment the Masked Autoencoder \cite{he2022masked} reconstruction loss, FLIP \cite{li2023scaling} randomly masks out input images to improve scaling, DACL \cite{verma2021towards} proposes a domain agnostic mixup strategy, SILC \cite{naeem2023silc} uses self-distillation, Mo \etal \cite{mo2024s} utilized specialist captions to generate pseudo labels for unpaired images, Zhang \etal \cite{zhang2023multi} propose Multi-task Paired Masking with Alignment to improve cross-modal interaction. Similarly there have been works that have identified the need to make the CLIP model focus on sub-regions in order to enhance its utility and downstream performance, GLoRIA \cite{huang2021gloria} considers the loss from local regions from within the image and reports, Alpha-CLIP \cite{sun2023alpha} uses the alpha channel to guide the CLIP model to focus on different regions of the image and generate the masks for all the images in the corpus using an image segmentation pipeline and TIER \cite{palepu2023tier} uses a regularization term to improve the local focus of the model. 

\noindent \textbf{Multi-modal Contrastive Learning in Medical Imaging:} Zhang \etal \cite{zhang2022contrastive} demonstrated enhanced downstream performance by jointly using chest X-ray and report pairing for training a contrastive learning model. This was further improved by Huang \etal \cite{huang2021gloria}, by exploiting local and global features from both modalities; Wang \etal \cite{wang2022medclip} and You \etal \cite{you2023cxr} achieved impressive results by using a Swin Tiny model as the image encoder and by adding modifications to the contrastive loss. Several other works have developed similar contrastive learning foundation models while utilizing the biomedical image texts to achieve impressive results \cite{huang2023visual, xu2023elixr, moor2023med, wu2023medklip, sowrirajan2010moco, tu2024towards}. Karargyris \etal \cite{karargyris2021creation} and Bigolin \etal \cite{bigolin2022reflacx} augment a subset of the MIMIC-CXR \cite{johnson2019mimic} samples with high quality eye-tracking and verbal transcripts from several radiologists. van Sonsbeek \etal \cite{van2023probabilistic} and Wang \etal \cite{wang2024gazegnn} utilize the heatmaps from eye-gaze to improve image classification.

\section{Method}

\noindent \textbf{Notations.} For a given chest X-ray image $I_i$ and its radiology report $T_i$, indexed by $i$ in our dataset, we denote their L2-normalized embeddings as $v_i$ and $t_i$ respectively, residing in a d-dimensional space ($v_i, t_i \in \mathcal{R}^d$). Image embeddings are obtained through an encoder $v_i = f(I_i)$ and text embeddings via $t_i = g(T_i)$. Applying an expert heatmap $E_i$ to an image results in the corresponding image embedding $v_i^E$. We denote the loss value for the $i$-th sample as $\mathcal{L}_i$. For the case of contrastive loss, this is computed in terms of some similarity measure between the embeddings, $sim(v_i, t_i)$, typically cosine similarity defined as $v_i \cdot t_i$.

\subsection{Background}
\label{sec: background}

Central to CLIP's effectiveness is the InfoNCE loss \cite{oord2018representation}, a mechanism engineered to optimize the similarity measures between corresponding (positive) pairs and to minimize those among non-corresponding (negative) pairs. The formulation of the CLIP loss objective is as follows: 

\begin{align}
    \mathcal{L}_{\text{text}} &= \mathop{\mathbb{E}}_{(t_i, v_i) \sim pos} \left[ - \log \frac{\exp(sim(t_i, v_i)/ \tau)}{\exp(sim(t_i, v_i)/\tau) + \sum_{j \neq i} \exp(sim(t_i, v_j)/\tau)}  \right] 
\end{align}

Total loss is then defined as, $\mathcal{L}_{\text{total}} = \frac{1}{2} \left( \mathcal{L}_{\text{text}} + \mathcal{L}_{\text{image}} \right)$, where $\mathcal{L}_{\text{image}}$ denotes the corresponding loss for the image to text mapping. Here, $\tau$ represents the temperature parameter that controls the scale of the similarity scores, typically framed as a learnable parameter during training. The loss expectation is taken over all the positive pairings in the dataset.

Theoretical results on CL indicate the concepts of \emph{alignment} and \emph{uniformity} as critical for the quality of embeddings \cite{wang2020understanding, wang2021understanding}. Alignment focuses on reducing the distance between positive pairs while uniformity seeks to evenly distribute the embeddings across the unit hypersphere, preventing extreme clustering that could impair the model's generalizability and discriminative capabilities. The alignment and uniformity can be defined formally as follows \cite{oh2024geodesic}:

\begin{align}
    \text{Alignment} &= - \mathbb{E}_{(v_i, t_i) \sim pos} \left[ \Vert v_i - t_i \Vert_2^2 - \min_{j  \neq i} \Vert v_i - t_j \Vert_2^2 \right] \\
    \text{Uniformity} &= - \log \mathbb{E}_{(v_i, t_j) \sim \mathcal{D}} \left[ \exp(-2 \Vert v_i - t_j \Vert_2^2) \right]
\end{align}

\begin{figure}[t]
  \centering
    \includegraphics[width=0.75\linewidth]{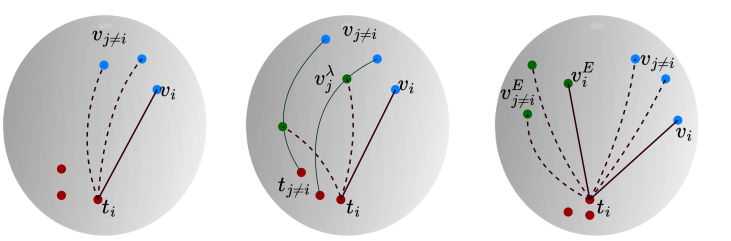}
  \caption{\textbf{Comparing eCLIP with $m^2$-mixup\cite{oh2024geodesic}.} \textit{(left)} Standard CLIP showing image-text positive pairs $(v_i, t_i)$ (solid line), while the other image embeddings serve as negative pairs (dashed line). \textit{(center)} the $m^2$-mixup creates negative pairs $(v^{\lambda}_j, t_i)$ via interpolation between embeddings along the geodesic. \textit{(right)} eCLIP adds expert image embedding, $v^E_i$, in addition to $v_i$ for text $t_i$, forming additional positive and negative pairs}
  \label{fig:compare-mixups}

\end{figure}
High intra-modal similarity, e.g. between two images as seen in \cref{fig:clip-cosine-similarities}, can inadvertently enhance similarity among negative pairs, inflating the denominator of the loss function in Equation (1), and, consequently, hurting the model's ability to differentiate between positive and negative pairs during training. A conventional method to counter this involves incorporating hard negative pairs, a strategy Oh \etal \cite{oh2024geodesic} employ by mixing embeddings from different modalities. While effective, this cross-modal mixup may obscure the semantic clarity of embeddings. As an alternative we propose increasing the dataset with additional positive pairs that exhibit minimal semantic overlap by integrating expert annotations. \cref{fig:compare-mixups} compares the positive and negative pair creation in eCLIP with traditional CLIP and $m^2$-mixup.

\subsection{Introducing Expert Annotations to CLIP}

\begin{algorithm}
\caption{eCLIP Algorithm}
\label{alg:eclip}
\begin{minipage}[t]{0.53\linewidth}
    \begin{algorithmic}[1]
        \REQUIRE Image Encoder $f(.)$
        \REQUIRE Text Encoder $g(.)$
        \STATE $\mathcal{L}_{\text{priming}} \gets 0$; $n_{\text{p}} \gets 0$
        \FOR{minibatch $\{x_i\}_{i=1}^{N}$}
            \STATE Unpack $x_i$ to $(T_i, I_i)$ and optionally $E_i$
            \STATE $t_i \gets g(T_i)$
            \STATE $v_i \gets f(I_i)$
            \STATE $p_{\text{uni}} \sim \text{Uniform}(0, 1)$
            \IF{$p_{\text{uni}} < p_{\text{curr}}$ \AND $E_i$ is provided}
                \STATE // Process expert image
                \STATE $\lambda \sim \text{Beta}(\alpha, \alpha)$
                \STATE $I_i^E \gets \text{HeatmapProcessor}(I_i, E_i)$
                \STATE $I_i^{\lambda} \gets I_i  \lambda + I_i^E  (1 - \lambda)$
                \STATE $v_i^{\lambda} \gets f(I_i^{\lambda})$
            \ENDIF
            \IF{$E_i$ is entirely ones}
                \STATE $I_i^{\text{R}} \gets \text{HeatmapProcessor}(I_i, E_i)$
                \STATE $\mathcal{L}_{\text{priming}} \gets \mathcal{L}_{\text{priming}} +  (I_i - I_i^{\text{R}})^2$
                \STATE $n_{\text{p}} \gets n_{\text{p}} + 1$
            \ENDIF            
        \ENDFOR
    \end{algorithmic}
\end{minipage}%
\hfill
\begin{minipage}[t]{0.43\linewidth}
    \flushleft
    \begin{algorithmic}
    \STATE // Compute Total Loss 
    \REQUIRE Temperature $\tau$
    \STATE $V \gets \text{List of } v_i \text{ for all } i$
    \STATE $T \gets \text{List of } t_i \text{ for all } i$
    \FOR{$i = 1$ \TO $N$}
        % \STATE // Append expert embeds
        \IF{$v_i^{\lambda}$ exists}
            \STATE $V \gets \text{Append}(V, v_i^{\lambda})$
            \STATE $T \gets \text{Append}(T, t_i)$
        \ENDIF
    \ENDFOR
    \STATE $\mathcal{L}_{\text{clip}} \gets \text{ClipLoss}(V, T, \tau)$
    \IF{$n_{\text{p}} > 0$}
        \STATE $\mathcal{L}_{\text{priming}} \gets \frac{1}{n_{\text{p}}}\mathcal{L}_{\text{priming}}$
    \ENDIF
    \STATE $\mathcal{L}_{\text{total}} \gets (1 - w_{p}) \cdot \mathcal{L}_{\text{clip}} + w_{p} \cdot \mathcal{L}_{\text{priming}}$ 
    
    \vspace{3pt}
    \textbf{Hyperparameters:}
    \vspace{-5pt}
    \begin{itemize}
        \item Batch size $N$
        \item Mixup Alpha $\alpha$
        \item Curriculum Prob. $p_{\text{curr}}$
        \item MSE Loss weight $w_{\text{p}}$
    \end{itemize}    

    \end{algorithmic}
\end{minipage}
\end{algorithm}

Our objective with eCLIP is to enhance the CLIP framework by integrating expert annotations -- radiologist eye-gaze heatmaps -- to diversify the pool of positive samples. The eCLIP model is designed to be compatible across all CLIP variants without modifying its core architecture. The \textbf{heatmap processor} (\cref{fig:main-figure}, \textit{right}) first converts the images and heatmaps into a sequence of patches and applies multi-headed attention (MHA) over the sequences. The patchified heatmap overlaid images serve as queries, while the original image's patches act as keys and values. The processed output is then reconstructed back to its original image format, enabling the standard CLIP image encoder to obtain expert image embeddings. These new embeddings and their text embedding pair introduce additional positive samples for the contrastive loss objective (\cref{fig:compare-mixups}, \textit{right}). 

However, the size of the expert annotated data is orders of magnitude smaller than the data available for CLIP training. To effectively leverage the scarce expert-annotated data, we implement \textbf{mixup} augmentation \cite{zhang2017mixup}. As illustrated in \cref{fig:main-figure} \textit{(left)}, this involves blending an original image $I_i$ with its expert version $I_i^E$ to create $I_i^{\lambda} = \lambda I_i + (1 - \lambda) I_i^E$, where $\lambda \sim \text{Beta}(\alpha, \alpha)$. (We set $\alpha = 0.3$ in all our experiments.) The eCLIP image encoder then processes $I_i^{\lambda}$ to produce the image embedding $v_i^{\lambda} = f(I_i^{\lambda})$. These expert embeddings form new positive pairs $(v_i^{\lambda}, t_i)$ as well as corresponding negative pairs, which are added with existing pairs $(v_i, t_i)$ during the computation of the CLIP InfoNCE Loss, $\mathcal{L}_i$. 

To seamlessly integrate expert annotations without disrupting the foundational training of the eCLIP model, we employ a phased \textbf{curriculum learning} strategy \cite{bengio2009curriculum}. This approach comprises a cold start phase where the model is initially trained without the expert annotations to establish a robust baseline. This phase accounts for about 10\% of the total training iterations. It then transitions into a warmup phase, gradually increasing the inclusion of expert examples from 0.05 to 0.5 probability over the next 30\% of iterations. Finally, a cooldown phase reduces expert example probability to 0.1 for the subsequent 40\% of iterations, fine-tuning the model's performance by balancing foundational and expert-driven insights. 

Additionally, we regularize the heatmap processor to behave as an identity function in scenarios where the heatmap is entirely composed of ones. We achieve this through a \textbf{priming} phase that coincides with the curriculum learning's cold start phase. We setup an auxillary mean-squared error loss to force the heatmap processor to reconstruct the original image $I_i$ when the heatmap $E_i = 1$. This priming ensures the heatmap processor's adaptability, allowing it to process expert annotations effectively when available, while falling back to the model's original performance in their absence. The total loss during this phase is, $\mathcal{L}_{\text{total}} = w_{\text{p}} \cdot \mathcal{L}_{\text{priming}} + (1 - w_{\text{p}}) \cdot \mathcal{L}_{\text{clip}}$, where $w_{\text{p}}$ is a hyperparameter which we set to 0.1. The pseudocode for eCLIP is shown in \cref{alg:eclip}.

\section{Experiments}
\label{sec:experiments}

Our experiments are designed to evaluate the influence of expert heatmap annotations on the quality of the learned representations. Unless stated otherwise, we assume that a large set of image-text pairs, of which a small fraction is annotated with eye-gaze heatmaps, is used for training, but for test samples no annotations are used. We utilize both quantitative measures and qualitative assessments to study the contributions of these annotations towards enhancing model performance. The source code is \href{https://github.com/ykumards/eCLIP}{available online}.

\subsection{Setup}
\label{experiment: setup}

\subsubsection{Baselines}
To validate our approach, we compare eCLIP against a model trained using traditional CLIP (referred to as the base model) and a ``naive'' baseline, where the expert annotated samples are directly added to the training set without using mixup or curriculum learning. We also examine the impact of two mixup methods: Domain Agnostic Contrastive Learning (DACL) \cite{verma2021towards} and $m^3$-mixup \cite{oh2024geodesic} which blends image and text embeddings to improve alignment and uniformity across modalities. While DACL is integrated during pretraining, $m^3$-mixup is applied post-pretraining, in a manner akin to fine-tuning. eCLIP can be applied to any variant of CLIP, which we demonstrate also with GLoRIA \cite{huang2021gloria} which has a Resnet50 image encoder. We introduce \textbf{two variants} of our technique: eCLIP, which integrates expert annotations during the initial CLIP pretraining phase, and eCLIP$^\mathcal{P}$, which instead continually finetunes a trained CLIP model with expert annotations, similar to $m^3$-mixup.

\subsubsection{Datasets}
\label{experiment: datasets}

For pretraining phase we utilize the \textbf{MIMIC-CXR} dataset \cite{johnson2019mimic}, which pairs roughly 200K chest X-rays with free-text radiology reports. The images were processed into JPEG format as described in \cite{johnson2019mimicjpg}, and the accompanying reports were stripped of unnecessary punctuation and tokenized using the Wordpiece scheme \cite{devlin2018bert}. We obtain the eye-gaze heatmap from the EGD-CXR dataset \cite{demner2016preparing} and process the eye-tracking data to obtain the normalized eye-gaze heatmap which are available for 1080 datapoints. 

Our evaluation setup includes multiple publicly available chest X-ray datasets, specifically \textbf{CheXpert} \cite{irvin2019chexpert}, \textbf{RSNA Pneumonia} \cite{shih2019augmenting}, \textbf{NIH CXR} \cite{wang2017chestx} and \textbf{Open-I} \cite{demner2016preparing}, each offering a distinct set of imaging and reporting characteristics. Following previous works \cite{wang2022medclip,huang2021gloria,you2023cxr}, we prepare the test sets from MIMIC and CheXpert, selecting 200 random samples for five specific pathologies from the CheXpert competition, resulting in 1000 samples for each dataset (MIMIC 5x200 and CheXpert 5x200, respectively). For the NIH-CXR dataset, we assembled a subset of 100 samples for each of 14 abnormalities, thereby creating CXR 14x100 test set. The Open-I dataset is utilized for text retrieval and radiology report generation tasks. For linear probe evaluations, we use CXR-8 \cite{wang2017chestx}, RSNA dataset and construct an OpenI-5 dataset by extracting labels from the `Problems' field within the reports that match the CheXpert competition labels.

\subsubsection{Training}
\label{experiment: setup-train}

We employed CLIP pretraining on the MIMIC-CXR dataset, utilizing a subset with roughly 1000 images with expert eye-gaze heatmap annotations, while validation and all other downstream evaluations proceed without these annotations. Our model architecture includes Swin Tiny, following recent studies\cite{wang2022medclip, you2023cxr}, alongside Vision Transformer (ViT) Small and Base, with image encoders pretrained on ImageNet and Clinical BERT \cite{alsentzer2019publicly} with max length of sequences set to 256 as the text encoder. We cropped images to $(224, 224)$ using random resized crop augmentation and turned off all other image augmentations. Pretraining utilized 8 AMD MI250X GPUs, maintaining an effective batch size of 512 for 10,000 steps. The learning rate was $1e^{-4}$ for standard CLIP, increased to $2e^{-4}$ for eCLIP variants, with cosine annealing plus a linear warmup for the first 10\% of iterations, weight decay of $1e^{-3}$, and learnable temperature parameter in the contrastive loss initialized at 0.07. Models for $m^3$-mixup and eCLIP$^\mathcal{P}$ are initialized with weights from CLIP pretraining and further finetuned for 1,000 iterations with a learning rate of $1e^{-5}$. Detailed setup information is available in the Supplement.

\subsection{Zero-shot Image Classification}

\setlength{\tabcolsep}{5pt}
\begin{table}[t]
    \centering
    \caption{Zero-shot classification performance on 4 X-ray datasets and model configurations, reported as macro-averaged F1 scores from three independent random seeds. The highest score per dataset and model configuration is underlined. The overall best-performing model for each dataset is highlighted in bold.}
    \label{tab:image-zero-shot}
    \begin{tabular}{@{\hspace{0.1cm}}l@{\hspace{0.2cm}}c@{\hspace{0.2cm}}c@{\hspace{0.2cm}}c@{\hspace{0.2cm}}c@{\hspace{0.1cm}}}
        \toprule
        \textbf{Model} & \multicolumn{4}{c}{\textbf{Dataset}} \\
        \midrule
        & Chexpert 5x200 & MIMIC 5x200 & RSNA & CXR 14x100 \\
        \midrule

        GLoRIA$_{\text{Resnet50}}$ 
        & $0.478_{\pm .023}$ & $0.457_{\pm .016}$ & $0.736_{\pm .024}$ & $0.155_{\pm .001}$ \\
        \quad $_{+ \text{naive}}$ 
        & $0.391_{\pm .069}$ & $0.334_{\pm .057}$ & $0.731_{\pm .023}$ & $0.113_{\pm .024}$ \\
        \quad $_{+ \text{DACL}}$ 
        & $0.506_{\pm .029}$ & $0.430_{\pm .011}$& $0.736_{\pm .007}$ & $0.158_{\pm .004}$ \\
        \quad $_{+ \text{$m^3$-mix}}$ 
        & \underline{$0.512_{\pm .005}$} & $0.467_{\pm .004}$ & $0.760_{\pm .006}$ & \underline{$0.160_{\pm .003}$} \\
        \quad $_{+ \text{expert (ours)}}$ 
        & $0.507_{\pm .004}$ & $0.430_{\pm .009}$ & $0.761_{\pm .017}$ & $0.156_{\pm .023}$ \\
        \quad $_{+ \text{expert}^{\mathcal{P}} \text{(ours)}}$ 
        & $0.507_{\pm .005}$ & \underline{$0.475_{\pm .008}$} & \underline{$0.775_{\pm .001}$} & $0.159_{\pm .001}$ \\
        \midrule

        CLIP$_{\text{Swin Tiny}}$ 
        & $0.517_{\pm .025}$ & $0.452_{\pm .002}$ & $0.808_{\pm .000}$ & $0.169_{\pm .003}$ \\
        \quad $_{+ \text{naive}}$ 
        & $0.532_{\pm .010}$ & $0.452_{\pm .022}$ & $0.807_{\pm .007}$ & $0.167_{\pm .007}$ \\
        \quad $_{+ \text{DACL}}$ 
        & $0.465_{\pm .008}$ & $0.389_{\pm .015}$ & $0.768_{\pm .018}$ & $0.101_{\pm .013}$ \\
        \quad $_{+ \text{$m^3$-mix}}$ 
        & $0.554_{\pm .006}$ & \underline{$0.469_{\pm .008}$} & $0.802_{\pm .004}$ & $0.179_{\pm .008}$ \\
        \quad $_{+ \text{expert (ours)}}$ 
        & $0.549_{\pm .016}$ & $0.445_{\pm .021}$ & $0.818_{\pm .004}$ & $0.172_{\pm .006}$ \\
        \quad $_{+ \text{expert}^{\mathcal{P}} \text{(ours)}}$
        & \underline{$0.558_{\pm .004}$} & $0.463_{\pm .007}$ & \underline{\boldmath$0.819_{\pm .000}$} & \underline{$0.192_{\pm .003}$} \\
        \midrule

        CLIP$_{\text{ViT Small}}$ 
        & $0.525_{\pm .024}$ & $0.441_{\pm .006}$ & $0.807_{\pm .006}$ & $0.159_{\pm .007}$\\
        \quad $_{+ \text{naive}}$ 
        & $0.534_{\pm .016}$ & $0.440_{\pm .019}$ & $0.805_{\pm .004}$ & $0.156_{\pm .017}$ \\
        \quad $_{+ \text{DACL}}$ 
        & $0.475_{\pm .025}$ & $0.398_{\pm 015}$ & $0.761_{\pm .009}$ & $0.133_{\pm .007}$ \\
        \quad $_{+ \text{$m^3$-mix}}$ 
        & $0.557_{\pm .002}$& \underline{$0.454_{\pm .003}$} & $0.809_{\pm .002}$ & $0.164_{\pm .002}$ \\
        \quad $_{+ \text{expert (ours)}}$ 
        & $0.545_{\pm .016}$ & $0.452_{\pm .013}$ & $0.803_{\pm .003}$ & $0.165_{\pm .007}$ \\
        \quad $_{+ \text{expert}^{\mathcal{P}} \text{(ours)}}$ 
        & \underline{$0.559_{\pm .001}$} & $0.439_{\pm .004}$ & \underline{$0.817_{\pm .001}$} & \underline{$0.165_{\pm .004}$} \\        
        \midrule        

        CLIP$_{\text{ViT Base}}$ 
        & $0.540_{\pm .017}$ & $0.465_{\pm .004}$ & $0.805_{\pm .001}$ & $0.183_{\pm .011}$ \\
        \quad $_{+ \text{naive}}$ 
        & $0.506_{\pm .011}$ & $0.426_{\pm .006}$ & $0.805_{\pm .004}$ & $0.151_{\pm .009}$ \\
        \quad $_{+ \text{DACL}}$ 
        & $0.474_{\pm .007}$ & $0.400_{\pm .002}$ & $0.759_{\pm .001}$ & $0.106_{\pm .003}$\\
        \quad $_{+ \text{$m^3$-mix}}$ 
        & $0.542_{\pm .021}$ & $0.465_{\pm .013}$ & $0.798_{\pm .004}$ & $0.183_{\pm .020}$ \\
        \quad $_{+ \text{expert (ours)}}$ 
        & \underline{\boldmath$0.563_{\pm .021}$} & \underline{\boldmath$0.477_{\pm .004}$} & \underline{$0.814_{\pm .003}$} & \underline{\boldmath$0.193_{\pm .017}$} \\
        \quad $_{+ \text{expert}^{\mathcal{P}} \text{(ours)}}$
        & $0.549_{\pm .003}$ & $0.452_{\pm .007}$ & $0.810_{\pm .001}$ & $0.185_{\pm .002}$ \\

        \bottomrule
    \end{tabular}
% \vspace{-10pt}
\end{table}

Following CLIP\cite{radford2021learning}, our zero-shot classification method categorizes images into predefined classes without direct finetuning, thus relying on the quality of embeddings generated during pretraining. To formulate the embedding of each class label, we first generate descriptive prompts to obtain a list of text embeddings corresponding to the label using the text encoder \cite{huang2021gloria, wang2022medclip, you2023cxr}. The mean of these embeddings is taken as the representation of the label. For each image, classification is then performed by matching the image embedding with its closest label embedding through cosine similarity. More details of the prompts used are provided in the Supplement.

\cref{tab:image-zero-shot} illustrates the zero-shot classification performance on CheXpert 5x200, MIMIC 5x200, RSNA, and CXR 14x100 datasets. The results, based on macro-averaged F1 scores from three random initializations, highlight eCLIP variants' superior performance over the base models across all datasets. While the $m^3$-mixup excels in MIMIC for certain architectures, eCLIP variants show broader generalization. Notably, eCLIP's advantages are more pronounced in multi-class scenarios (CheXpert, MIMIC and CXR14) compared to binary classification on RSNA.

\subsection{Sample Efficiency}
\begin{figure}[t]
% \vspace{-10pt}
    \centering
    \includegraphics[width=0.8\textwidth]{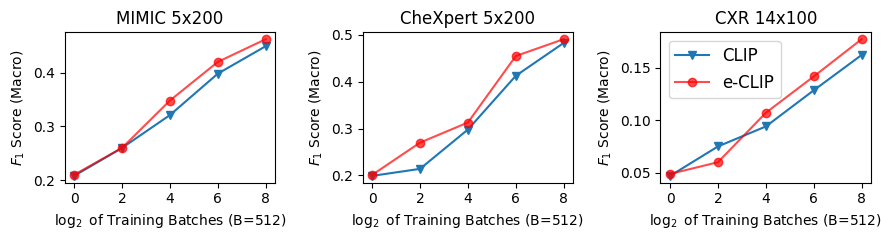}

    \centering
    \includegraphics[width=0.8\textwidth]{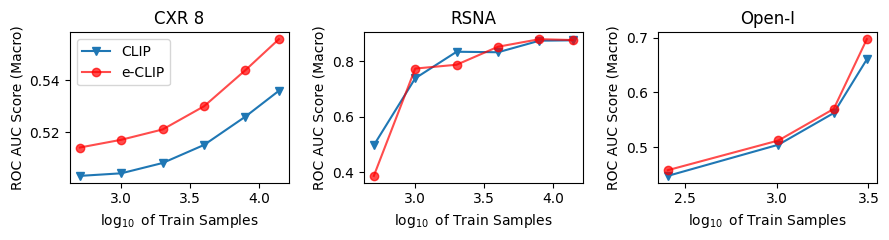}
    
    \caption{\textbf{Sample Efficiency.} \textit{(top row)} Zero-shot performance on three multi-label classification test sets for CLIP and eCLIP Swin Tiny models, trained with varying amounts of training batches. \textit{(bottom row)} Linear probe scores with varying amounts of training data.} 
    \label{fig:sample_efficiency}
% \vspace{-10pt}
\end{figure}

Sample efficiency measures how well a model learns from limited amount of training data. eCLIP improves this efficiency by using expert annotated images to form new positive and negative pairs, aiming to improve the quality of the learned embeddings. To test this, we first looked at zero-shot classification performance, adjusting the number of training batches available for pretraining. Results, shown in \cref{fig:sample_efficiency} \textit{(top row)}, reveal that eCLIP is more sample efficient across MIMIC 5x200, CheXpert 5x200 and CXR 14x100 datasets compared to the base model. Additionally, by applying supervised fine-tuning (SFT) with a linear probe on class-imbalanced datasets -- CXR-8, RSNA and OpenI-5 (\cref{experiment: datasets}) -- eCLIP demonstrates stronger performance in multi-label classification tasks, CXR-8 and OpenI-5 and remains competitive in binary classification for RSNA. This is shown \cref{fig:sample_efficiency} \textit{(bottom row)} where we plot the ROC AUC scores against different training sample sizes for linear probing. These findings highlight eCLIP's ability to effectively learn from fewer samples.

\subsection{Text retrieval and retrieval augmented generation (RAG)}

\begin{table}[t]
    \centering
    \begin{minipage}[t]{0.48\textwidth}
        \caption{Performance on Image to Report retrieval with Open-I dataset. We report the Recall@\{1, 5, 10\}}
        \label{tab:image-text-retrieval}
        \begin{tabular}{@{}lccc@{}}
            \toprule
            \textbf{Model}                               & R@1 & R@5 & R@10  \\ \midrule
            
            CLIP$_{\text{Swin Tiny}}$                   &  3.1  & 7.6   & 11.3  \\
            \quad $_{+ \text{$m^3$-mix}}$              &  2.4  & 6.5   & 9.8  \\
            \quad $_{+ \text{expert (ours)}}$          &  \underline{3.7}  & \underline{9.4}   & \underline{13.4}  \\ 
            \quad $_{+ \text{expert}^{\mathcal{P}} \text{(ours)}}$
            &  3.1  & 8.2   & 11.7  \\ \midrule             
    
            CLIP$_{\text{ViT Base}}$                    &  3.7  & 9.2   & 13.2  \\
            \quad $_{+ \text{$m^3$-mix}}$              &  4.1  & 8.8   & 13.0  \\
            \quad $_{+ \text{expert (ours)}}$          &  \underline{\textbf{4.4}}  & \underline{\textbf{10.3}}  & \underline{\textbf{13.5}}  \\
            \quad $_{+ \text{expert}^{\mathcal{P}} \text{(ours)}}$     
            &  3.9  & 9.4  & 13.2  \\ 
            
            \bottomrule
        \end{tabular}

        \vspace{10pt}

        \caption{Performance on report generation with Open-I dataset. We report the BLEU-2 score (BL-2), BERT recall score\cite{zhang2019bertscore} (B-R), Cosine similarity between the sentence embeddings of the generated and ground-truth report for MPNet \cite{song2020mpnet} Sentence Transformer model \cite{reimers2019sentence}  (S$_{emb}$), and for the CheXBERT model\cite{smit2020combining} (CB$_{emb}$)}
            \label{tab:radiology-report}
            \begin{tabular}{@{\hspace{0.1cm}}lllll@{\hspace{0.cm}}}
            \toprule
            \textbf{Model}                             & BL-2   & B-R               & S$_{emb}$         & CB$_{emb}$  \\ \midrule
            CLIP                                       & 0.172  &  \textbf{0.713}   & 0.791             & 0.492   \\
            \scriptsize{+ \text{$m^3$-mix}}              & 0.172  &  0.711            & 0.788             & 0.496   \\
            eCLIP                                      & \textbf{0.177} &  0.712    & \textbf{0.795}    & \textbf{0.506} \\
        \bottomrule
        \end{tabular}        
        
    \end{minipage}\hfill
    \begin{minipage}[t]{0.48\textwidth}
        \caption{\textbf{Ablation Study with Swin Tiny.} Zero-shot performance on CheXpert (CXP) and CXR14 (C14) datasets for the base CLIP and models with expert annotation integration (+E) is presented. Methods include mask multiplication ($\odot$), CNN, and Multi-headed Attention (MHA) encoders. Key augmentations: Mixup (+M), Curriculum Learning (+C), and Encoder Priming (+P) demonstrate performance gains. A control with a random mask (\textit{rand}) confirms the significance of expert annotations. We report macro-averaged F1 scores from three random initialization} 
        \label{tab:ablation}
        \begin{tabular}{@{}lcc}
            \toprule
            \textbf{Method} & CXP & C14 \\ \midrule
            
            \scriptsize{Base}                            
            & \scriptsize{$0.517_{\pm .024}$}   & \scriptsize{$0.169_{\pm .003}$} \\
            
            \scriptsize{$\odot$ Mask} \tiny{(+ E)}       
            & \scriptsize{$0.540_{\pm .019}$}      & \scriptsize{$0.165_{\pm .006}$}\\
            
            \scriptsize{CNN Encoder} \tiny{(+ E)}        
            & \scriptsize{$0.534_{\pm .012}$} & \scriptsize{$0.163_{\pm .008}$}  \\
            
            \scriptsize{MHA Encoder} \tiny{(+ E)}        
            & \scriptsize{$0.534_{\pm .013}$} & \scriptsize{$0.153_{\pm .002}$}  \\
            
            \begin{tabular}[c]{@{}l@{}}\scriptsize{MHA Encoder}\\ \quad \tiny{(+ E, M)}\end{tabular} 
            & \scriptsize{$0.532_{\pm .018}$} & \scriptsize{$0.160_{\pm .010}$} \\
            
            \begin{tabular}[c]{@{}l@{}}\scriptsize{MHA Encoder}\\ \quad \tiny{(+ E, M, C)}\end{tabular} 
            & \scriptsize{$0.545_{\pm .008}$} & \scriptsize{$0.173_{\pm .018}$} \\
            
            \begin{tabular}[c]{@{}l@{}}\scriptsize{MHA Encoder}\\ \quad \tiny{(+ rand, M, C, P)}\end{tabular} 
            & \scriptsize{$0.537_{\pm .020}$} & \scriptsize{$0.166_{\pm .003}$} \\
            
            \begin{tabular}[c]{@{}l@{}}\scriptsize{MHA Encoder}\\ \quad \tiny{(+ E, M, C, P)}\end{tabular} 
            & \scriptsize{$0.549_{\pm .016}$} & \scriptsize{$0.172_{\pm .006}$} \\            
            
            \bottomrule
        \end{tabular}
    \end{minipage}
% \vspace{-10pt}
\end{table}

\begin{table}[t]
\centering
\caption{\textbf{Random samples of generated report.} For each image in the Open-I dataset, the five closest text snippets based on embedding cosine similarity is used as prompts for  Mistral 7B LLM. Utilizing in-context learning, we prompt the LLM with two such snippet-report pairs. The conditions that the generated report identified correctly are highlighted in \textcolor{highlightGreen}{green} while those it missed are shown in \textcolor{highlightRed}{red}.}
\begin{tabular}{@{}p{2cm}p{4.5cm}p{4.5cm}@{}} 
\toprule
\textbf{Problems} & \textbf{Ground Truth} & \textbf{Generated} \\ 
\midrule
\scriptsize{normal} 
& \scriptsize{The lungs and pleural spaces show no acute abnormality. Heart size and pulmonary vascularity within normal limits}.
& \scriptsize{The lungs are clear with \textcolor{highlightGreen}{no evidence of consolidation, pneumothorax, or pleural effusion. Heart size and mediastinal contour are normal.} \textcolor{highlightGreen}{No acute abnormalities} noted in the visualized osseous structures.} \\ \midrule

\scriptsize{Pulmonary Atelectasis. Pleural Effusion. Cardiomegaly. Pulmonary Congestion}
& \scriptsize{AP and lateral views were obtained. Bibasilar atelectasis and small left-sided pleural effusion. Stable cardiomegaly. No pneumothorax. \textcolor{highlightRed}{Mild pulmonary vascular congestion.}}
& \scriptsize{The lungs show \textcolor{highlightGreen}{bibasilar airspace opacities}, possibly due to infection or \textcolor{highlightGreen}{atelectasis. Heart size is mildly enlarged. No pneumothorax} \textcolor{highlightRed}{or pleural effusion}.} \\ \midrule

\scriptsize{Lung. Density. Emphysema}
& \scriptsize{There is hyperexpansion. The heart size is normal. There is no pleural effusion or pneumothorax. Two circular densities overlying the right ribs which were not present in the XXXX CT. No focal infiltrates}
& \scriptsize{The lungs are \textcolor{highlightGreen}{hyperexpanded with clear fields and no acute consolidations or pleural effusions. Heart size and pulmonary vascularity are within normal limits.} \textcolor{highlightRed}{Degenerative changes are present in the thoracic spine. A calcified granuloma is noted in the left perihilar region.} No pneumothorax or large pleural effusion is observed.} \\ 
\bottomrule
\end{tabular}
\label{tab: report-gen-samples}
% \vspace{-10pt}
\end{table}

To compare eCLIP's cross-modal functionality with that of CLIP we focused on text retrieval task using the Open-I dataset, which consists of pairs of X-rays and radiology reports. We used the FAISS vector database \cite{douze2024faiss} to index the text embeddings generated by the text encoder. For a given X-ray image $I_i$, we then retrieve the closest text reports from the database based on the cosine similarity in the embedding space, $min_{j} (v_i \cdot t_j)$. Results in \cref{tab:image-text-retrieval} compare the performance of eCLIP against CLIP in text retrieval measured in Recall@1, 5, and 10. The performance of eCLIP indicates a notable improvement in its embedding quality. Note that our evaluation followed a strict criterion for recall computation, where a retrieval was counted as successful only if the exact correct report was identified. While more nuanced measures based on semantic similarity could be employed \cite{you2023cxr}, we opted this approach to maintain a clear and simple evaluation framework.

Next we extend our analysis from retrieval to report generation using a frozen Large Language Model (LLM), Mistral 7B Instruct v2 \cite{jiang2023mistral}, aiming to generate radiology reports through Retrieval Augmented Generation (RAG). This setup tests the CLIP model's capacity to retrieve texts which can be used to prompt an LLM to generate a report without finetuning on medical data. First we randomly selected 389 samples from the Open-I dataset for testing and utilized the FAISS database to index the reports from the remaining samples (i.e., training set). Given a test image we retrieve five closest reports from the training set and use them in the prompt for the LLM to generate a report for the test image. The eCLIP variant showed a small but consistent improvement over the base model in generating reports, as indicated in \cref{tab:radiology-report}. A comparative analysis of generated report versus ground truth shown in \cref{tab: report-gen-samples}, with discrepancies marked, further validates the effectiveness of eCLIP's embeddings in supporting complex cross-modal tasks. Additional details, including LLM prompts and generated report samples are available in the Supplement.

\subsection{Embedding Quality}

\begin{figure}[t]
    \centering
    \begin{subfigure}[t]{0.9\textwidth}
        \includegraphics[width=\textwidth]{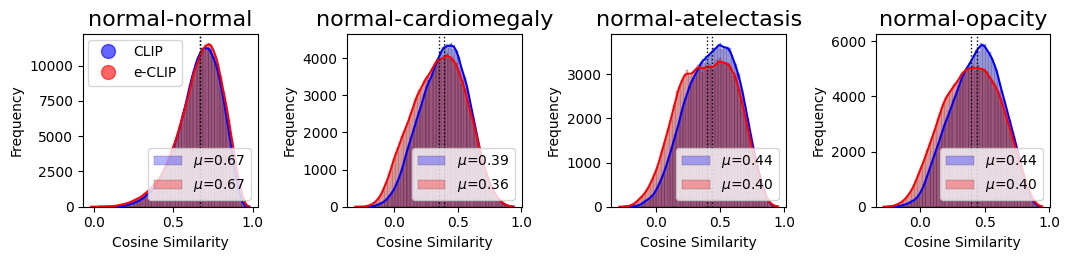}
    \end{subfigure}
    \begin{subfigure}[t]{0.9\textwidth}
        \includegraphics[width=\textwidth]{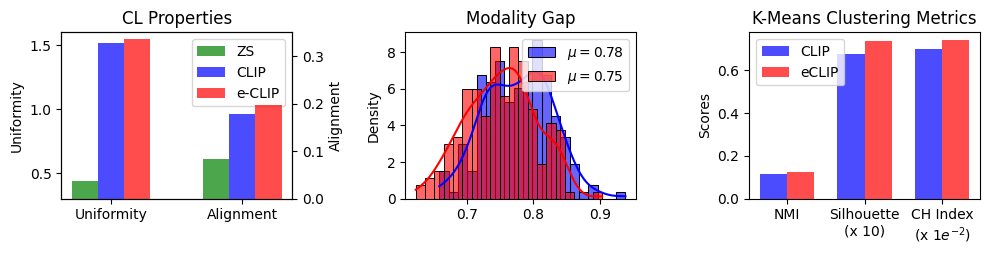}
    \end{subfigure}
    \caption{\textbf{Qualitative Analysis of CLIP Pretraining.} \textit{top row} illustrates the cosine similarity distributions for CLIP and eCLIP image embeddings. \textit{bottom left and center} sections display uniformity, alignment, and modality gap comparisons among the internet pretrained model (ZS), CLIP pretrained on MIMIC, and eCLIP. \textit{bottom right} details K-means clustering metrics for image embeddings with k=5 for both CLIP and eCLIP models.}
    \label{fig:embedding-quality}
% \vspace{-10pt}
\end{figure}

For qualitative evaluations, we first examine the histogram of the cosine similarities of the embeddings from different abnormality subgroups obtained from the CLIP image encoder. In \cref{fig:embedding-quality} \textit{(top row)}, we can see that the similarities for the CLIP model has considerably dropped below 1 after continual pretraining on MIMIC-CXR compared to \cref{fig:clip-cosine-similarities}. This indicates that the model's ability to distinguish between different conditions has improved. The introduction of expert annotations in the eCLIP variant further improves this with mean cosine similarities for `normal' versus cardiomegaly, atelectasis and opacity dropping to 0.36, 0.4 and 0.4 respectively. 

Our evaluation of uniformity and alignment reveals that eCLIP surpasses both the MIMIC-pretrained and internet-pretrained models in these key metrics, indicating a marked improvement in the quality of embeddings (\cref{fig:embedding-quality}, \textit{bottom left}). We also note a modest decrease in the modality gap with eCLIP (\cref{fig:embedding-quality}, bottom center). Clustering analysis via K-means (with k=5 for 5 abnormalities in data) highlights eCLIP's superior performance in grouping abnormalities, as seen from improved scored in Normalized Mutual Information (NMI), Silhouette score, and Calinski-Harabasz (CH) index (\cref{fig:embedding-quality}, bottom right). 

\subsection{Ablation Study}

Our ablation study with the Swin Tiny encoder shows the impact of key components in our eCLIP model: multi-headed attention (MHA) layer for heatmap procesing, curriculum learning for phased introduction of expert annotations, mixup augmentation to compensate for limited number of export annotated data and priming of heatmap processor during initial training phase. Results shown in \cref{tab:ablation} reveal that the MHA-based heatmap processor improves zero-shot classification performance on CheXpert 5x200 and CXR 14x100 datasets compared to basic methods like direct application of heatmaps as mask ($\odot Mask)$ or using a CNN encoder. We note a significant performance drop with randomly generated heatmaps versus expert eye-gaze heatmaps. This highlights that while our methodological improvements contribute to the performance gains, the integration of meaningful, expert-derived signals is essential for achieving optimal results.

\section{Conclusion}
We introduce eCLIP, an adaptation of CLIP, demonstrating the integration of radiologist eye-tracking heatmap to overcome challenges faced in multi-modal contrastive learning. This study highlights the impact of integrating these high-quality expert annotations on improving the quality of learned embeddings and assess its influence on sample efficiency and cross-modal retrieval tasks. An important future research direction would be extending this approach to include expert annotations from the text modality (e.g., by adapting SimCSE \cite{gao2021simcse}) and to leverage the temporal dynamics of eye-tracking data by aligning the sequential frames with the corresponding report snippets.

\subsubsection{Limitations}
Our study is limited by the small size of expert annotated data and thus does not comprehensively analyze the impact of size or distribution of expert annotations across different abnormalities. eCLIP also incurs extra computational costs during training due to the additional forward pass required for processing expert images in the warmup and cool-down phases. Additionally, the clinical relevance of generated radiology reports has not been validated by medical experts, relying instead on standard metrics known for potential biases and inaccuracies in reflecting clinical accuracy\cite{yu2023evaluating}.

\section*{Acknowledgements}
This work was supported by the Research Council of Finland (Flagship programme: Finnish Center for Artificial Intelligence FCAI, and grants 352986, 358246) and EU (H2020 grant 101016775 and NextGenerationEU). We acknowledge the computational resources provided by Aalto Science-IT project. We acknowledge CSC for awarding this project access to the LUMI supercomputer, owned by the EuroHPC Joint Undertaking, hosted by CSC (Finland) and the LUMI consortium through Finland.

\bibliographystyle{splncs04}
\bibliography{main}

\title{Supplement}

\titlerunning{eCLIP}

\author{Yogesh Kumar\orcidlink{0000-0002-7961-8596} \and
Pekka Marttinen\orcidlink{0000-0001-7078-7927}}

\authorrunning{Y.~Kumar and P.~Marttinen}

\institute{Department of Computer Science, Aalto University, Finland}

\maketitle

\appendix
\section{Detailed Experiment Setup}

\subsection{Pretraining with Expert Annotations}

For pretraining both CLIP and eCLIP models, we utilize the MIMIC-CXR dataset. Expert annotations, in the form of heatmaps, are derived from a subset of MIMIC-CXR, the EGD-CXR dataset \cite{karargyris2021creation}, which comprises of 1080 samples. We employ the author's official preprocessing code to convert the eye-tracking fixation data into heatmaps. The data tuple (image, text, heatmap) is then used for training with contrastive learning. We employ two dataloader: one for the main dataset without heatmap (``main loader'') and another for the subset with expert heatmaps (``expert loader''). In each training iteration, one batch is fetched from each loader; the CLIP model processed only the main batch, while the eCLIP model has the flexibility to use either just the main batch or both batches. \cref{lst:pseudo-code} shows the Pytorch-like pseudocode for the eCLIP model.

The utilization of the expert batch in eCLIP is determined by a curriculum probability, initially set to zero during the cold start phase. This probability linearly increases to $p_{max}$ during the warmup phase, then linearly decreases to $p_{min}$ during the cool-down phase, where it remains for the remainder of the training process. $p_{max}$ was set to 0.5 for all experiments, while $p_{min}$ was set to 0.05 for the ViT Base model and to 0.1 for all other models. 

\begin{figure}[ht]
\begin{lstlisting}[language=Python, caption=PyTorch-like pseudocode for eCLIP implementation, label=lst:pseudo-code]
def heatmap_processor(image, heatmap):
    # reshapes and transposes have been omitted for brevity
    pathches = patchify(image)
    heatmap_patches = patchify(heatmap * image) 
    processed_patches = multi_head_attention(
        q=heatmap_patches, k=patches, v=patches)
    reconstructed_image = unpatchify(processed_patches)
    return reconstructed_image

class ExpertClipImageEncoder:
    self.base = ViT()
    self.projector = ProjectionBlock()

    def forward(image, heatmap=None):
        mse_loss = None
        B, C, H, W = image.size()
        # priming 
        reconstructed_image = heatmap_processor(
        image, torch.ones((B, 1, H, W)))
        mse_loss = mse_loss_fn(image, reconstructed_image)
        
        image_features = self.base(image)
        # project to shared embedding dimension
        image_embed = self.projector(image_features)

        if heatmap is not None:
            expert_image = heatmap_processor(image, heatmap)
            # mixup augmentation
            lambda_ = beta(alpha=0.3).sample()
            expert_image_m = mixup(image, expert_image, lambda_)
    
            expert_image_features = self.base(expert_image_m)
            expert_image_embed = self.projector(expert_image_features)

        return image_embed, expert_image_embed, mse_loss
        
# include expert batch based on curriculum learning probability
# curriculum_prob is varied between cpmin to cpmax (typically, 0.1 & 0.5)
use_expert =  np.random.rand() < curriculum_prob

# compute clip loss
clip_loss = clip_loss_fn(
    concat([image_embed, expert_image_embed]) if use_expert else image_embed, 
    concat([text_embed, text_embed]) if use_expert else text_embed
)
\end{lstlisting}
\end{figure}

\subsection{$m^2$-mixup vs $m^3$-mixup}

We illustrate $m^2$-mixup in Fig. \ref{fig:compare-mixups}, where embeddings from image and text domains are mixed to mitigate the modality gap, as proposed by Oh et al. \cite{oh2024geodesic}. Oh et al. \cite{oh2024geodesic} further introduce $m^3$-mixup, which combines $m^2$-mixup with corresponding unimodal mixups. Specifically,

\[
    \mathcal{L}_{m^3} = \mathcal{L}_{\text{CLIP}} + \mathcal{L}_{m^2} + \mathcal{L}_{uni}
\]

For further details, please refer to Oh et al. \cite{oh2024geodesic}.

\subsection{Linear Probe Experiments}
In our linear probe experiments, we utilize the CLIP and eCLIP with Swin Tiny as the image encoder following other recent similar works \cite{wang2022medclip, you2023cxr}. The pretrained model's image encoder is entirely frozen and we append a linear layer for classification. This layer's output dimension is set to 1 for the Pneumonia dataset, 8 for CXR-8 and 5 for OpenI-5. We allocate 10\% of the training data as validation set and conduct training over 5 epochs with a cosine decay learning rate schedule with linear warmup for 10\% of the total training steps. Base learning rates are set to $2e^{-5}$ for the Pneumonia dataset and $1e^{-5}$ for both CXR-8 and OpenI-5. We employ binary cross entropy as the loss function and the model selection for testing is based on the epoch with the lowest validation loss.

\subsection{Zero-shot Classification}
Following the CLIP paper\cite{radford2021learning}, we generate descriptive prompts for each label, mirroring the patterns found in the radiology reports of our pretraining data. For example, within the pneumonia detection task, a `normal' X-ray is prompted as ``Chest radiograph with normal findings, no signs of pneumonia'', while a prompt for pneumonia diagnosed X-rays would read ``Radiograph of the chest displaying multifocal opacities, suggestive of viral pneumonia''. We apply the ensemble promoting technique, where we generate multiple variations of each label's prompt to create a list of text embeddings for each label. The mean of these embeddings serves as the representation for the corresponding label. The specific prompts utilized for each label have been included in the Supplement section. 

These prompts are converted into embeddings using the text encoder of the trained CLIP model, while the images are processed using the corresponding image encoder to produce image embeddings. Classification is then performed by selecting the label whose text emebdding is most similar to the image embedding, as determined by cosine similarity. We use the prompts used in \cite{huang2021gloria}, samples from which are shown below.

\newpage
\begin{mdframed}[backgroundcolor=gray!10,linewidth=2pt]
    \textbf{ZS Prompts } \\

    \noindent Atelectasis - mild subsegmental atelectasis \\
    Cardiomegaly - cardiac silhouette size is mildly enlarged \\
    Consolidation - increased reticular consolidation at the lower lung zone \\
    Edema - mild pulmonary edema \\
    Pleural Effusion - stable right bilateral pleural effusion \\
    Pneumonia - Bronchopneumonia pattern suggestive of bacterial infection \\ 
\end{mdframed}

\newpage
\section{Additional Results}

In Table \ref{tab:image-zero-shot-acc} we show the zero-shot classification accuracies for the CLIP, eCLIP and baseline models. 

\setlength{\tabcolsep}{5pt}
\begin{table}
    \centering
    \caption{Zero-shot classification performance on 4 X-ray datasets and model configurations, reported as accuracy from three independent random seeds. The highest score per dataset and model configuration is underlined. The overall best-performing model for each dataset is highlighted in bold.}
    \label{tab:image-zero-shot-acc}
    \begin{tabular}{@{\hspace{0.1cm}}l@{\hspace{0.2cm}}c@{\hspace{0.2cm}}c@{\hspace{0.2cm}}c@{\hspace{0.2cm}}c@{\hspace{0.1cm}}}
        \toprule
        \textbf{Model} & \multicolumn{4}{c}{\textbf{Dataset}} \\
        \midrule
        & Chexpert 5x200 & MIMIC 5x200 & RSNA & CXR 14x100 \\
        \midrule

        GLoRIA$_{\text{Resnet50}}$ 
        & $0.498_{\pm .017}$ & $0.462_{\pm .014}$ & $0.731_{\pm .013}$ & $0.173_{\pm .002}$ \\
        \quad $_{+ \text{naive}}$ 
        & $0.409_{\pm .061}$ & $0.369_{\pm .044}$ & $0.669_{\pm .049}$ & $0.145_{\pm .020}$ \\
        \quad $_{+ \text{DACL}}$ 
        & \underline{$0.530_{\pm .016}$} & $0.438_{\pm .007}$& $0.752_{\pm .007}$ & $0.179_{\pm .008}$ \\
        \quad $_{+ \text{$m^3$-mix}}$ 
        & $0.525_{\pm .004}$ & $0.469_{\pm .004}$ & $0.748_{\pm .004}$ & $0.171_{\pm .003}$ \\
        \quad $_{+ \text{expert (ours)}}$ 
        & $0.518_{\pm .001}$ & $0.436_{\pm .008}$ & $0.730_{\pm .023}$ & $0.179_{\pm .028}$ \\
        \quad $_{+ \text{expert}^{\mathcal{P}} \text{(ours)}}$ 
        & $0.520_{\pm .004}$ & \underline{$0.478_{\pm .008}$} & \underline{$0.753_{\pm .000}$} & $0.168_{\pm .001}$ \\
        \midrule

        CLIP$_{\text{Swin Tiny}}$ 
        & $0.529_{\pm .020}$ & $0.454_{\pm .003}$ & $0.799_{\pm .004}$ & $0.188_{\pm .006}$ \\
        \quad $_{+ \text{naive}}$ 
        & $0.536_{\pm .008}$ & $0.459_{\pm .018}$ & $0.793_{\pm .002}$ & $0.188_{\pm .012}$ \\
        \quad $_{+ \text{DACL}}$ 
        & $0.481_{\pm .005}$ & $0.398_{\pm .020}$ & $0.758_{\pm .005}$ & $0.124_{\pm .019}$ \\
        \quad $_{+ \text{$m^3$-mix}}$ 
        & $0.561_{\pm .006}$ & \underline{$0.467_{\pm .005}$} & $0.802_{\pm .001}$ & $0.208_{\pm .007}$ \\
        \quad $_{+ \text{expert (ours)}}$ 
        & $0.549_{\pm .016}$ & $0.443_{\pm .023}$ & $0.799_{\pm .004}$ & $0.195_{\pm .003}$ \\
        \quad $_{+ \text{expert}^{\mathcal{P}} \text{(ours)}}$
        & \underline{$0.565_{\pm .004}$} & $0.465_{\pm .006}$ & \underline{\boldmath$0.810_{\pm .001}$} & \underline{\boldmath$0.210_{\pm .001}$} \\
        \midrule

        CLIP$_{\text{ViT Small}}$ 
        & $0.524_{\pm .023}$ & $0.441_{\pm .007}$ & $0.796_{\pm .000}$ & $0.179_{\pm .007}$\\
        \quad $_{+ \text{naive}}$ 
        & $0.532_{\pm .017}$ & $0.454_{\pm .028}$ & $0.793_{\pm .003}$ & $0.169_{\pm .022}$ \\
        \quad $_{+ \text{DACL}}$ 
        & $0.485_{\pm .024}$ & $0.402_{\pm 015}$ & $0.753_{\pm .009}$ & $0.154_{\pm .007}$ \\
        \quad $_{+ \text{$m^3$-mix}}$ 
        & \underline{$0.561_{\pm .002}$}& $0.455_{\pm .003}$ & $0.786_{\pm .002}$ & $0.182_{\pm .001}$ \\
        \quad $_{+ \text{expert (ours)}}$ 
        & $0.548_{\pm .017}$ & \underline{$0.456_{\pm .010}$} & $0.795_{\pm .005}$ & \underline{$0.185_{\pm .015}$} \\
        \quad $_{+ \text{expert}^{\mathcal{P}} \text{(ours)}}$ 
        & $0.556_{\pm .001}$ & $0.437_{\pm .004}$ & \underline{$0.805_{\pm .001}$} & $0.180_{\pm .003}$ \\        
        \midrule        

        CLIP$_{\text{ViT Base}}$ 
        & $0.540_{\pm .011}$ & $0.470_{\pm .007}$ & $0.788_{\pm .012}$ & $0.200_{\pm .006}$ \\
        \quad $_{+ \text{naive}}$ 
        & $0.503_{\pm .010}$ & $0.434_{\pm .011}$ & $0.787_{\pm .002}$ & $0.176_{\pm .008}$ \\
        \quad $_{+ \text{DACL}}$ 
        & $0.484_{\pm .006}$ & $0.406_{\pm .002}$ & $0.737_{\pm .001}$ & $0.204_{\pm .098}$\\
        \quad $_{+ \text{$m^3$-mix}}$ 
        & $0.544_{\pm .015}$ & $0.458_{\pm .005}$ & $0.771_{\pm .009}$ & $0.188_{\pm .005}$ \\
        \quad $_{+ \text{expert (ours)}}$ 
        & \underline{\boldmath$0.564_{\pm .023}$} & \underline{\boldmath$0.476_{\pm .005}$} & \underline{$0.799_{\pm .003}$} & $0.204_{\pm .020}$ \\
        \quad $_{+ \text{expert}^{\mathcal{P}} \text{(ours)}}$
        & $0.557_{\pm .011}$ & $0.464_{\pm .018}$ & $0.782_{\pm .010}$ & $0.197_{\pm .0.013}$ \\

        \bottomrule
    \end{tabular}
\end{table}

\newpage 

\begin{figure}[t]
    \centering
    \includegraphics[width=0.8\textwidth]{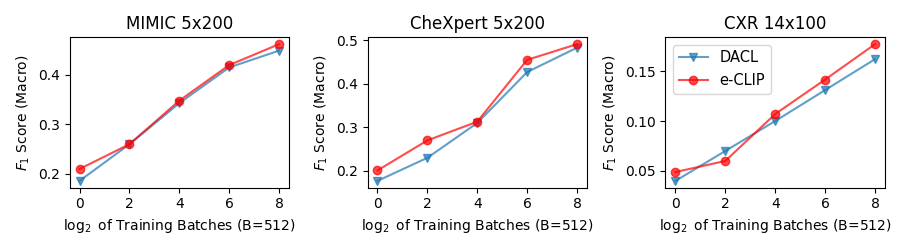}

    \centering
    \includegraphics[width=0.8\textwidth]{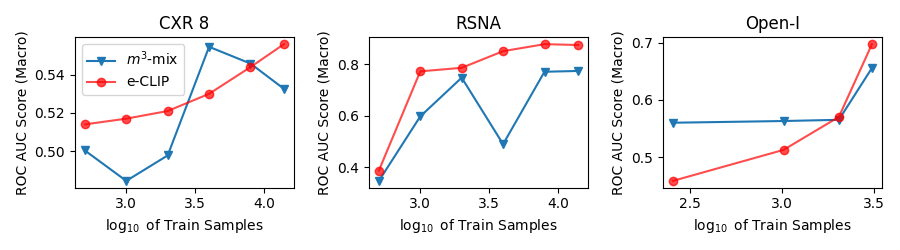}
    
    \caption{\textbf{Sample Efficiency.} \textit{(top row)} Zero-shot performance on three multi-label classification test sets for DACL and eCLIP Swin Tiny models, trained with varying amounts of training batches. \textit{(bottom row)} Linear probe scores with varying amounts of training data for $m^3$-mixup and eCLIP Swin Tiny models. } 
    \label{fig:sample_efficiency-app}
\end{figure}

Figure \ref{fig:sample_efficiency-app} demonstrates the sample efficiency of different models. The top row shows the zero-shot performance on three multi-label classification test sets for DACL and eCLIP Swin Tiny models, trained with varying amounts of training batches. This highlights each model’s ability to generalize with limited data. The bottom row presents the linear probe scores for m3-mixup and eCLIP Swin Tiny models, evaluated with different amounts of training data. This illustrates how quickly each model learns and performs as more data is provided. These results underscore the importance of sample efficiency in model performance.

\clearpage
\newpage
\section{Visualize Embeddings with UMAP}

\begin{figure}[h]
    \centering
    \includegraphics[width=0.6\textwidth]{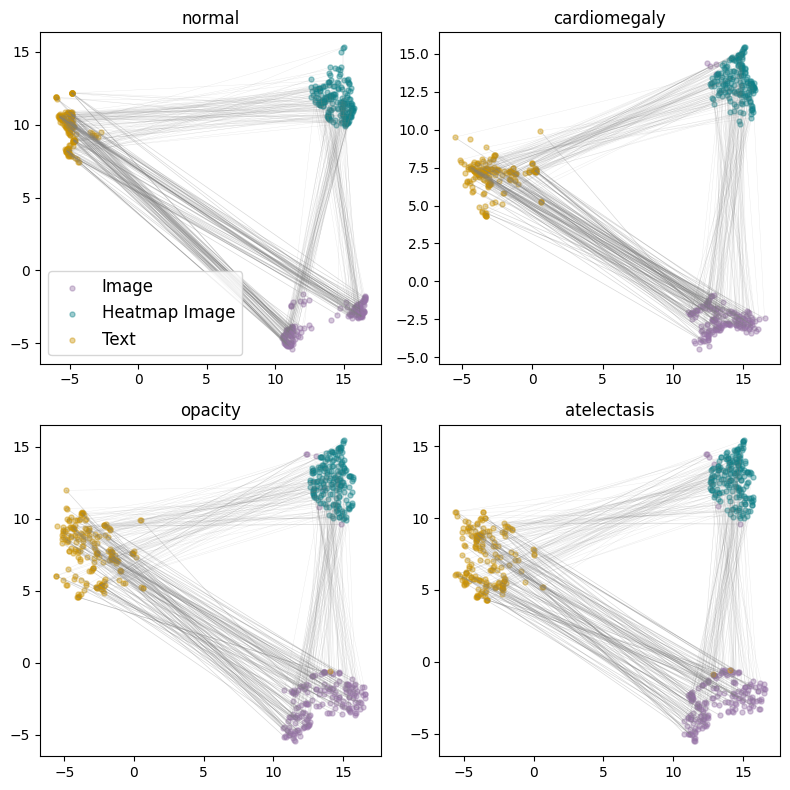}
    \caption{\textbf{2D UMAP Projection of Embeddings} Figure shows the UMAP projection of the Image, Text and heatmap processed Image embedding generated by eCLIP with Swin Tiny encoder. We use Open-I dataset for image and text and since expert annotation is unavailable for this dataset, we generate random uniform masks to simulate heatmaps. }
    \label{fig:masked-umap-all}
\end{figure}

We utilize the trained eCLIP model with the Swin Tiny encoder to examine UMAP projections of embeddings derived from the Open-I dataset. This dataset is categorized into subgroups based on the presence of specific abnormalities, as indicated in the `Problem' column, which contains radiologist annotations for each image. Four primary abnormalities --`normal,' `cardiomegaly,' `atelectasis,' and `opacity' -- form the basis of our subgroup categorization. We ensure that samples within each subgroup are mutually exclusive, containing only one of these abnormalities.

To generate embeddings, we use the trained image and text encoders from our eCLIP model. Since the Open-I dataset lacks actual expert-annotated heatmaps, we simulate this condition by creating random heatmaps for each image. Thus we obtain the standard image embedding ($v_i$), text embedding ($t_i$) and expert image embedding ($v_i^E)$ for each sample across the subgroups. These embeddings are projected into a 2D space using UMAP with cosine similarity as the metric. Subsequently, we visualize the 2D UMAP projections for each subgroup separately, facilitating a detailed inspection of the embedding distribution and the influence of expert annotations on the model's representation space. This is shown in \cref{fig:masked-umap-all}

\newpage
\section{Retrieval Augmented Generation of Radiologist Report}

\begin{figure}[h]
    \centering
    \includegraphics[scale=0.25]{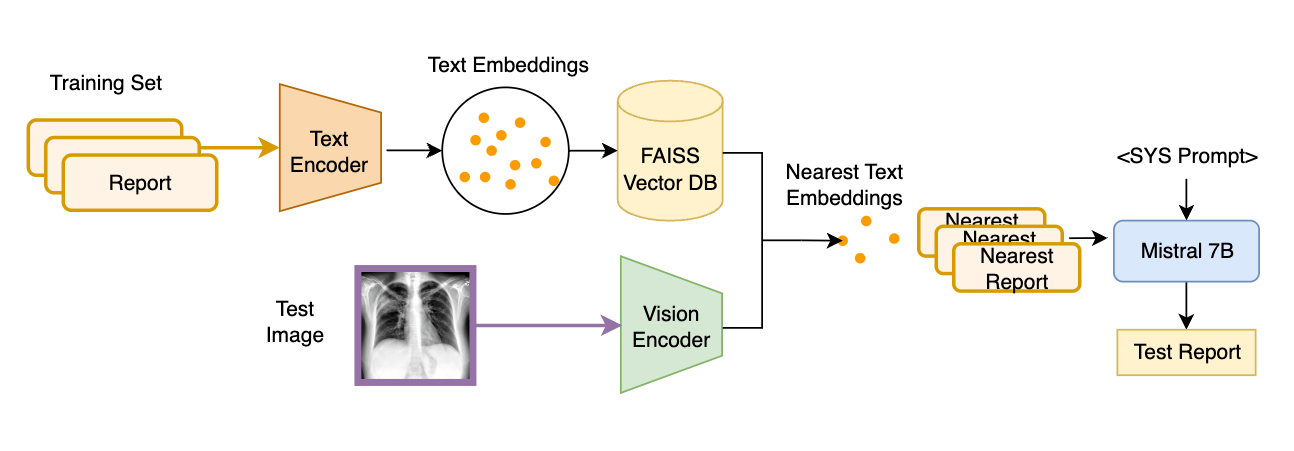}
    \caption{\textbf{Retrieval Augmented Generation of Radiology Reports.} Radiology reports from the training corpus are encoded using the CLIP/eCLIP's text encoder to obtain text embeddings, which are then stored in a FAISS vector database. For a test image, the corresponding image embedding is obtained using the CLIP/eCLIP's image encoder. This image embedding is queried against the FAISS database to find the nearest text embeddings, which are used as prompts for the Mistral 7B Large Language Model (LLM). The LLM then generates a test report based on these prompts.}
    \label{fig:llm-report-generation}
\end{figure}

We detail our approach to generate radiologist reports by augmenting a Large Language Model (LLM) with retrieved report snippets from the training corpus. This method is designed to tackle the challenge of generating medically relevant and coherent radiology report without direct image examination or explicit fine-tuning on medical datasets. The process involves the following key steps:

\begin{enumerate}
    \item \textbf{Text Embedding and Indexing:} We utilize the Open-I dataset to create the source of the radiology reports. These reports are processed through the CLIP/eCLIP trained text encoder to produce embeddings of dimension 512. These embeddings are then normalized and indexed using FAISS \cite{douze2024faiss} vector database, facilitating efficient retrieval based on similarity.

    \item \textbf{Text Retrieval and Clustering:} For a given test X-ray image, we first compute its embedding using the CLIP/eCLIP trained image encoder and then query the FAISS index to retrieve the closest report snippets. The ``closeness'' is based on cosine similarity of the normalized embeddings, ensuring that the retrieved texts are semantically relevant to the image's medical context. We use K-means to categorize the embeddings of these snippets into distinct groups. This clustering ensures the selection of representative sentences that encapsulate the primary observations within each group, thereby preserving the diversity of the retrieved reports. We require five closest snippets for prompting the LLM, so we retrieve four times this from the FAISS index for clustering.

    \item \textbf{Report Generation with LLM:} Leveraging the retrieved snippets as context, we employ a frozen LLM, Mistral 7B\cite{jiang2023mistral}, to generate a comprehensive radiology report. The aim is to produce reports that closely mimic those written by radiologists, based on the insights from the retrieved texts. 
\end{enumerate}

\cref{fig:llm-report-generation} shows the schematic of the steps involved in report generation using a frozen LLM.

\subsubsection{LLM Prompts}
We use the following system and user prompts to use the frozen Mistral 7B model to generate the radiology report. We provide two exemplars of   retrieved reports and the corresponding generated report as samples for the In-Context Learning (ICL) as the user prompt for the LLM.

\begin{mdframed}[backgroundcolor=gray!10,linewidth=2pt]
    \textbf{<SYSTEM PROMPT> } \\
    You are to act as a radiologist, trained to generate radiology reports. Your task is to synthesize the information from the closest report snippets provided below into a comprehensive and medically accurate radiologist report for each case. Craft a comprehensive response that is concise, succinct, and focuses on the key findings and potential diagnoses. Your report should maintain a professional tone, with clarity and precision in medical terminology, suitable for medical experts. Remember to be concise, succinct, and focus on the key findings and potential diagnoses, avoiding unnecessary elaboration.    
\end{mdframed}

\begin{mdframed}[backgroundcolor=gray!10,linewidth=2pt]
    \textbf{USER } \\
    The following snippets are from reports closely related to the patient's X-ray image. \\
    < Retrieved Text > \\
    Based on these, generate a radiologist report.
\end{mdframed}

\subsubsection{Evaluation Metrics} For evaluating the generated radiology reports, we employ metrics traditionally used in text generation and translation fields, namely BLEU-2 and BERT Score \cite{zhang2019bertscore}. We also compute the cosine similarity between the embeddings of the generated reports and the ground truth, as derived from a reference model. This approach allows for a broader assessment of semantic congruence. Specifically, we utilize sentence transformer models \cite{reimers2019sentence} known for their effectiveness in sentence-level comparison tasks. We employ the `all-mpnet-base-v2' \cite{song2020mpnet} model for its general semantic understanding, and the `CheXBERT' model \cite{smit2020combining}, for its domain-specific performance in medical classification tasks. These models facilitate a more comprehensive and contextually relevant evaluation of the linguistic and clinical content of the generated reports.

\subsubsection{Generated Radiology Report Samples} We provide more randomly sampled generated radiology reports in \cref{tab: more-report-gen-samples}

\begin{table}[t]
\centering
\caption{\textbf{More Random samples of generated report.} For each image in the Open-I dataset, the five closest text snippets based on embedding cosine similarity is used as prompts for  Mistral 7B LLM. Utilizing in-context learning, we prompt the LLM with two such snippet-report pairs.}
\begin{tabular}{@{}p{2cm}p{4.5cm}p{4.5cm}@{}} % Adjust the width as needed
\toprule
\textbf{Problems} & \textbf{Ground Truth} & \textbf{Generated} \\ 
\midrule
\scriptsize{normal} 
& \scriptsize{The heart is normal in size. The mediastinum is unremarkable. The lungs are clear.}.
& \scriptsize{The lungs are clear with no evidence of consolidation, pleural effusion, or pneumothorax. Heart size is within normal limits. The visualized osseous structures are unremarkable.} \\ \midrule

\scriptsize{normal}
& \scriptsize{XXXX XXXX and lateral chest examination was obtained. One AP view is expiratory and was repeated. The heart silhouette is normal in size and contour. Aortic XXXX appear unremarkable. Lungs demonstrate no focal infiltrates. There is no effusion or pneumothorax.}
& \scriptsize{The heart size is normal. The lungs are clear with no evidence of consolidation, pneumothorax, or pleural effusion. No acute abnormalities are noted in the visualized osseous structures. The previously seen right upper lobe mass lesion is not present on the current study.} \\ \midrule

\scriptsize{Pulmonary Atelectasis; Pleural Effusion; Cardiomegaly; Pulmonary Congestion}
& \scriptsize{AP and lateral views were obtained. Bibasilar atelectasis and small left-sided pleural effusion. Stable cardiomegaly. No pneumothorax. Mild pulmonary vascular congestion.}
& \scriptsize{The lungs show bibasilar airspace opacities, possibly due to infection or atelectasis. Heart size is mildly enlarged. No pneumothorax or pleural effusion.}  \\ \midrule

\scriptsize{normal}
& \scriptsize{The lungs and pleural spaces show no acute abnormality. Heart size and pulmonary vascularity within normal limits.}
& \scriptsize{The lungs are clear with no evidence of consolidation, pneumothorax, or pleural effusion. Heart size and mediastinal contour are normal. No acute abnormalities noted in the visualized osseous structures.}  \\ \midrule

\scriptsize{Lung; Density; Emphysema}
& \scriptsize{There is hyperexpansion. The heart size is normal. There is no pleural effusion or pneumothorax. Two circular densities overlying the right ribs which were not present in the XXXX CT. No focal infiltrates}
& \scriptsize{The lungs are hyperexpanded with clear fields and no acute consolidations or pleural effusions. Heart size and pulmonary vascularity are within normal limits. Degenerative changes are present in the thoracic spine. A calcified granuloma is noted in the left perihilar region. No pneumothorax or large pleural effusion is observed.}  \\ \midrule

\scriptsize{Lung; Spine}
& \scriptsize{Lungs are mildly hyperexpanded. The lungs are clear. There is no focal airspace consolidation. No pleural effusion or pneumothorax. Heart size and mediastinal contour are within normal limits. There are diffuse degenerative changes of the spine.}
& \scriptsize{The heart size and mediastinal contours are within normal limits. The lungs are clear, but with slightly low lung volumes and subtle left basal opacities. No pleural effusion, pneumothorax, or acute osseous abnormalities are noted. There is evidence of degenerative changes in the thoracic spine on the lateral view.}  \\ \midrule

\scriptsize{Calcified Granuloma; Opacity}
& \scriptsize{Heart size and pulmonary vascularity appear within normal limits. Calcified granuloma is present in the right base. No pneumothorax or pleural effusion is seen. In the lateral right base is identified an ill-defined somewhat oblong opacity. This was not present on the previous study. The remainder of the lungs appear clear.}
& \scriptsize{The lungs are clear with mild hyperexpansion, suggestive of underlying emphysema. Heart size is normal. No pleural effusions, pneumothoraces, or focal consolidations. The cardiomediastinal silhouette and osseous structures are unremarkable.}  \\ 

\bottomrule
\end{tabular}
\label{tab: more-report-gen-samples}
% \vspace{-10pt}
\end{table}
% \end{document}
\end{document}